\documentclass[conference]{IEEEtran}
\IEEEoverridecommandlockouts

\usepackage{graphicx}
\usepackage{booktabs}
\usepackage{array}
\usepackage{amsmath}
\usepackage{url}
\usepackage{float}
\usepackage[hidelinks]{hyperref}
\usepackage[backend=biber,style=ieee,sorting=none,maxnames=6,minnames=3]{biblatex}
\usepackage{microtype}
\usepackage{float}
\usepackage{eso-pic}
\usepackage{graphicx}
\usepackage{booktabs}
\begin{document}
\raggedbottom

\AddToShipoutPictureFG*{%
  \AtPageUpperLeft{%
    \raisebox{-12pt}[0pt][0pt]{%
      \makebox[\paperwidth][r]{%
        \parbox{0.57\paperwidth}{\raggedleft\scriptsize
        12th IEEE International Conference on Sustainable Technologies and Engineering (IEEE iCOSTE 2026)\\
        1-3 November 2026, Cox's Bazar, Bangladesh}%
        \hspace{0.25in}%
      }%
    }%
  }%
  \AtPageLowerLeft{%
    \raisebox{0.28in}[0pt][0pt]{%
      \hspace{0.62in}%
      \makebox[0pt][l]{\scriptsize 979-8-3195-2804-9/26/\$31.00 \copyright 2026 IEEE}%
    }%
  }%
}

\title{ThetreStageAI: Multi-View Retrieval-Augmented Generation for Structured Bengali Theatre Production Planning}

\makeatletter
\newcommand{\linebreakand}{%
  \end{@IEEEauthorhalign}
  \hfill\mbox{}\par
  \mbox{}\hfill\begin{@IEEEauthorhalign}
}
\makeatother

\author{
\IEEEauthorblockN{\hspace{-1.5cm}Md. Mehedi Hasan Naeem}
\IEEEauthorblockA{\textit{\hspace{-1.5cm}Computer Science and Engineering} \\
\textit{\hspace{-1.5cm}Jatiya Kabi Kazi Nazrul Islam University} \\
\hspace{-1cm}Mymensingh, Bangladesh \\
\hspace{-1.2cm}mehedinaeem00@gmail.com}
\and
\IEEEauthorblockN{Mst. Kamrunnahar Ruma}
\IEEEauthorblockA{\textit{Computer Science and Engineering} \\
\textit{Jatiya Kabi Kazi Nazrul Islam University} \\
Mymensingh, Bangladesh \\
kamrunnaharruma242@gmail.com}

\and
\IEEEauthorblockN{Shakila Sultana}
\IEEEauthorblockA{\textit{Computer Science and Engineering} \\
\textit{Jatiya Kabi Kazi Nazrul Islam University} \\
Mymensingh, Bangladesh \\
shakila161470@gmail.com}
\and
\IEEEauthorblockN{\hspace{-2cm}Nafiza Anjum}
\IEEEauthorblockA{\textit{\hspace{-1cm}Computer Science and Engineering} \\
\textit{\hspace{-1cm}Jatiya Kabi Kazi Nazrul Islam University} \\
\hspace{-0.5cm}Mymensingh, Bangladesh \\
\hspace{-0.5cm}nafizaanjum.2002@gmail.com}

\and
\IEEEauthorblockN{Nasreen Jannatun Nemu}
\IEEEauthorblockA{\textit{Computer Science and Engineering} \\
\textit{Jatiya Kabi Kazi Nazrul Islam University} \\
Mymensingh, Bangladesh \\
monalisafruti11223@gmail.com}
\and
\IEEEauthorblockN{Md. Sujan Ali}
\IEEEauthorblockA{\textit{Computer Science and Engineering} \\
\textit{Jatiya Kabi Kazi Nazrul Islam University} \\
Mymensingh, Bangladesh \\
sujan\_cse@jkkniu.edu.bd}
}

\maketitle

\begin{abstract}
Staging a theatre performance requires coordinating dialogue, actor movement, stage directions, lighting focus, cue timing, and equipment constraints within the same production plan. This work presents \emph{ThetreStageAI}, a retrieval augmented framework for structured Bengali theatre production planning. Rather than retrieving from one mixed corpus, the system maintains separate scene, actor-blocking, and lighting views. Bengali text is embedded with BGE-M3 and indexed in view-specific Qdrant collections. Retrieved references are combined with a production brief and passed to Gemini, which generates schema-constrained JSON containing characters, scenes, dialogue, stage directions, blocking, lighting, and optional sound cues. Structural and semantic checks are applied before acceptance, with at most one constrained repair attempt. We also construct a human-guided, LLM-assisted synthetic dataset of 500 short Bengali theatre records that expands to 1,500 retrieval documents. Dataset diagnostics confirm consistent identifiers and valid lighting fields while also revealing substantial dialogue repetition. A paired camera-ready evaluation compares single-combined and full multi-view retrieval under matched settings. The source code, dataset, and evaluation materials are publicly available.
\end{abstract}


\section{Introduction}
Writing the dialogue is only one part of preparing a stage production. During rehearsal, a director also has to decide where actors stand and move, when lighting changes, which fixture is used, and how those choices line up with the dramatic moment. Recent work has explored controllable drama through structured director and actor roles \cite{han2024ibsen}, interactive playwriting with language models \cite{rosa2021theaitre}, and computational support for theatrical lighting \cite{seo2025lighting}. These studies show that language models can assist creative work, but they do not remove the need to coordinate narrative and technical decisions.

RAG provides a practical way to give a language model task-specific material without retraining its parameters \cite{lewis2020rag}. For our use case, however, a single retrieval pool was too coarse. Dialogue examples, actor movement, and lighting design serve different purposes, even when they come from the same source script. We therefore designed ThetreStageAI around three retrieval views and combine their evidence only after view-specific search.

The sustainability relevance of this work lies in resource-aware production planning. The user brief limits generation to available lighting fixtures, and cue relationships can be checked before physical rehearsal. This may reduce avoidable trial-and-error setup and supports reuse of existing stage equipment. We do not claim measured energy or carbon savings because physical energy consumption was not evaluated in this study.

The paper makes these following contributions:
\begin{enumerate}
    \item a multi-view RAG architecture with separate scene, actor-blocking, and lighting retrieval while retaining source provenance;
    \item a human-guided, LLM-assisted synthetic Bengali theatre dataset with 500 source records and 1,500 retrieval-view documents;
    \item a schema-constrained generation layer that keeps dialogue, blocking, lighting, and optional sound information in one production object;
    \item a two-stage validation process that checks both data structure and production-brief constraints before an output is accepted; and
    \item an open-source prototype with retrieval traces, experiment logging, and a reproducible paired evaluation of retrieval configurations.
\end{enumerate}

\section{Related Work}
Prior work spans RAG, multilingual retrieval, drama generation, and computational stage design. ThetreStageAI brings these areas together by integrating narrative, actor blocking, and lighting cues in one structured production framework.

\subsection{Retrieval-Augmented Generation}
RAG links language-model generation with retrieved evidence rather than relying solely on parametric knowledge. Lewis et al. established the basic formulation for knowledge-intensive NLP tasks \cite{lewis2020rag}, and later surveys frame RAG as a pipeline spanning indexing, retrieval, context construction, generation, and evaluation \cite{fan2024survey}. Building on this foundation, Self-RAG lets the model decide when to retrieve and critique its own output \cite{asai2024selfrag}, while Corrective RAG adds a repair step for cases where the retrieved evidence is weak \cite{yan2024crag}. For evaluation, RAGAs and ARES both score retrieval and generation separately instead of relying on a single end-to-end metric \cite{es2024ragas, saadfalcon2024ares}. What reaches the generator also depends on how text is represented and indexed \cite{kamalipour2026vectors}. For this reason, we treat retrieval and validation as separate stages that can each be inspected, rather than as one opaque step.

\subsection{Dense Retrieval and Multilingual Embeddings}
Dense retrieval matters here because Bengali narrative text and technical cue descriptions are written in very different styles. Sentence-BERT gave an efficient way to compare sentences semantically \cite{reimers2019sbert}, and BGE-M3 extends dense retrieval across languages, retrieval functions and granularities \cite{chen2024m3}. That makes it a reasonable starting point for the mixed narrative and technical Bengali text in our three retrieval views.

\subsection{Language Models for Drama and Role-Based Generation}
Other work applies language models directly to drama and character behavior. THEaiTRE explored machine-assisted, interactive playwriting \cite{rosa2021theaitre}. IBSEN had directors and actors agents collaborate to generate controllable interactive drama scripts \cite{han2024ibsen} and ProTriPlay structured interactive theater around director, screenwriter and actor roles \cite{yu2026protriplay}. RoleLLM and TimeChara show that role consistency and temporally coherent character behavior must be evaluated explicitly, since general language ability does not guarantee them \cite{wang2024rolellm, ahn2024timechara}. Research on automatic story evaluation adds a similar caution: model-based judgments cannot fully replace human assessment \cite{chhun2024stories}.

\subsection{Computational Stage and Lighting Design}
Stage design has developed along a largely separate line of research. Music and speech emotion have been used to drive lighting behavior \cite{hsiao2017lighting, vryzas2017augmenting}, while visual objectives have supported exploratory lighting design \cite{shimizu2019visual}. More recent systems include StageDesigner for script-conditioned scenography \cite{gan2025stagedesigner}, LoRA-adapted LLMs for theatrical lighting specifications \cite{seo2025lighting}, Skip-BART and SeqLight for generating stage-light sequences \cite{zhao2025skipbart, zhao2026seqlight} and LightingGen for DMX-oriented entertainment lighting \cite{wang2026lightinggen}. Most of these works focus on a single production layer. ThetreStageAI instead brings narrative content, actor blocking, and lighting cues together in one event-aligned production object.

\section{Proposed ThetreStageAI Framework}
The proposed framework combines multi-view retrieval, structured generation, validation and coordinated theatre cues in a single production-planning pipeline. Its goal is to keep narrative content, actor movement and lighting information connected and verifiable.

\begin{figure}[H]
    \centering
    \includegraphics[width=0.94\columnwidth]{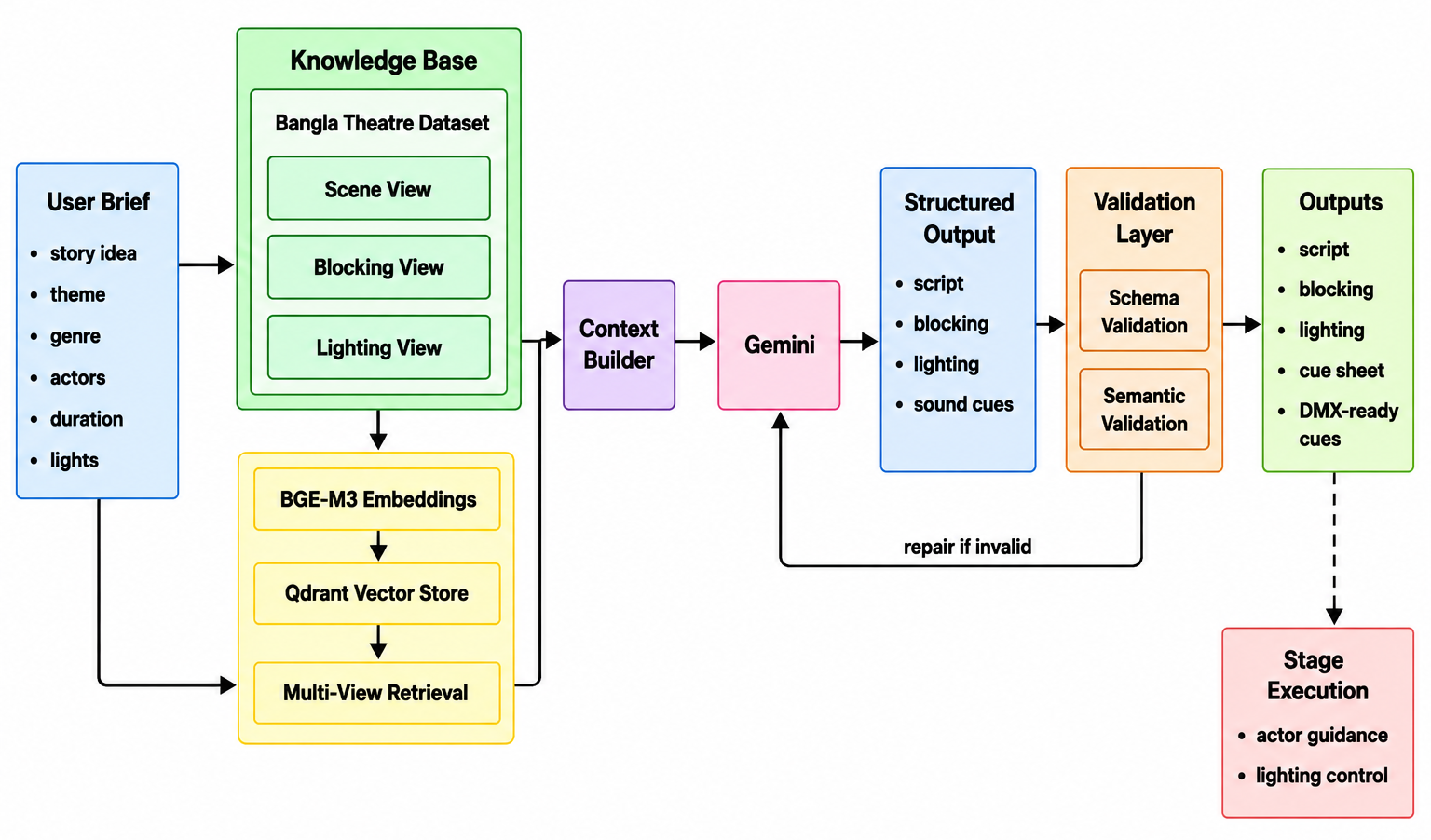}
    \caption{ThetreStageAI system architecture.}
    \label{fig:overall}
\end{figure}

Fig.~\ref{fig:overall} shows the implementation built for this work. A request starts with a production brief, which is used to retrieve relevant theatre material. The retrieved context is combined with the brief and passed to the language model, and the generated production is validated before it is accepted. Physical lighting control is deliberately kept outside this planning loop.

\subsection{Production Brief and Structured Objective}
Each request provides a story idea and production details: theme, genre, language, number of actors, target duration, stage size, available fixtures, scene time and desired emotion. Some of these fields shape the retrieval queries, while others are checked again after generation as explicit constraints. The output is a structured production object rather than free-form prose, containing a title, theme, genre, characters, and scenes, where each scene may hold dialogue, stage directions, blocking cues, lighting cues, and optional sound cues.

\subsection{Multi-View Retrieval}
Rather than indexing each theatre record as one long document, we split every source record into three retrieval views. The scene view keeps material useful for narrative generation, including summary, dialogue, and stage directions. The blocking view covers actor movement, triggers, and stage zones. The lighting view keeps fixture, focus, color, intensity, fade duration, and related movement context. Each view carries the original source identifier, searchable Bengali text, metadata, and a structured payload.

Only the searchable text is embedded with BGE-M3 \cite{chen2024m3}, and vectors with their payloads are stored in separate Qdrant collections, keeping the retrieval stages modular \cite{fan2024survey}. At generation time, the production brief is converted into view-specific queries. In the default full multi-view mode, the retriever returns three scene, two blocking, and two lighting references. Duplicate sources within a view are removed, context is bounded, and each selected item is logged with its source ID, rank, score, and view type.

\begin{figure}[H]
    \centering
    \includegraphics[width=0.92\columnwidth]{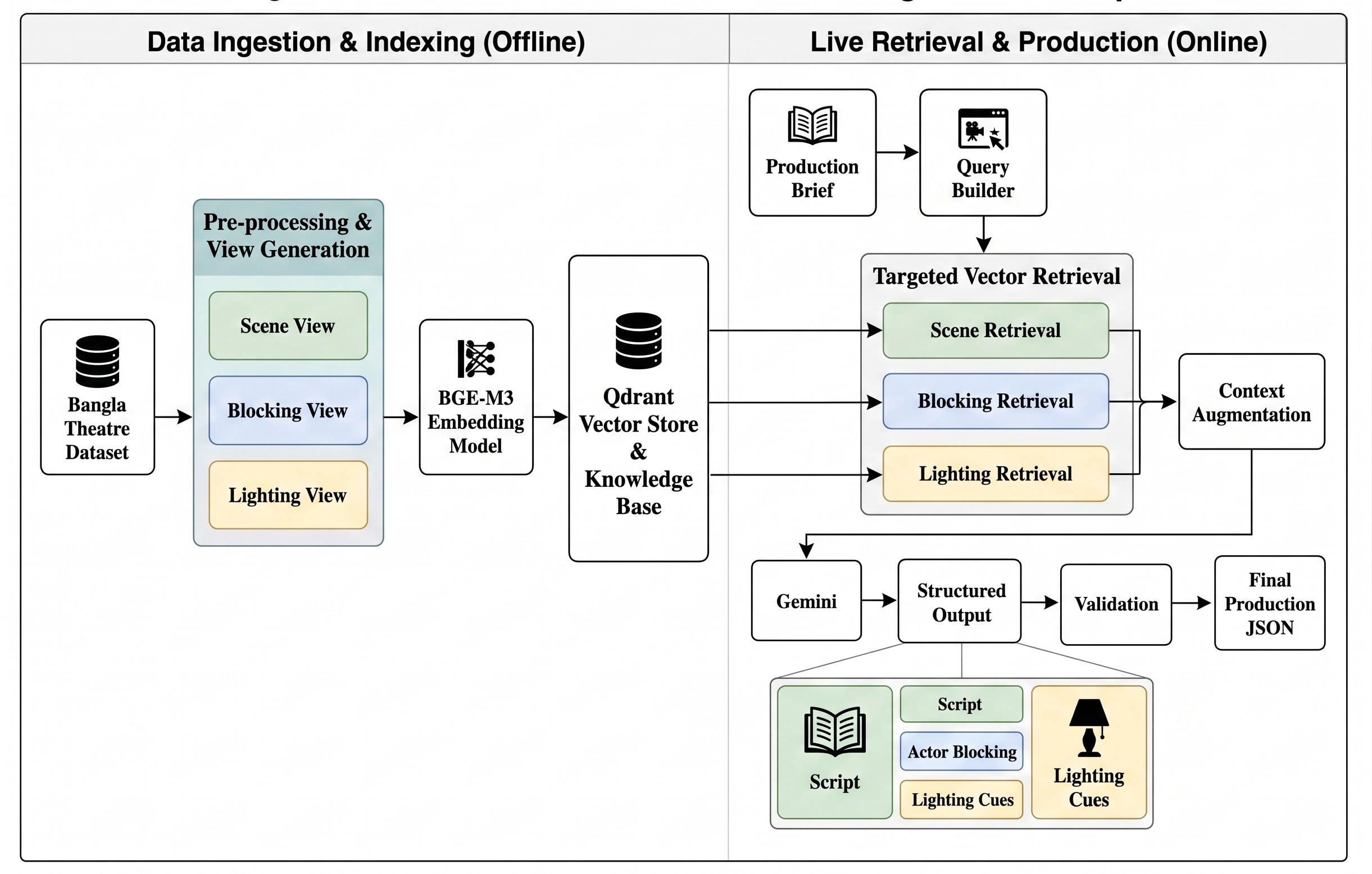}
    \caption{Multi-view retrieval and context construction.}
    \label{fig:rag}
\end{figure}

The separation in Fig.~\ref{fig:rag} is intentional. A scene may be close to the requested story but still be a poor example for actor movement or lighting. Searching the views independently keeps those differences visible rather than hiding them in one combined retrieval result.

\subsection{Gemini-Based Structured Generation}
After retrieval, Gemini receives the user brief, the selected theatre references and the production schema. We ask for JSON instead of free text because fluent language alone does not guarantee that an LLM will follow format and task constraints \cite{kumar2024llm}. We chose Gemini as the generation backend for its multilingual, general-purpose capabilities \cite{gemini2023}. It sits behind a provider interface, so retrieval and validation do not depend on anything specific to Gemini.

\subsection{Validation and Repair}
Gemini's output is not accepted right away. It first goes through a structural check, in which strict Pydantic models verify the required fields, stage-zone values, numeric ranges, unique identifiers, known speakers and actors, and whether cue triggers point to valid scene events or dialogue IDs. A second check compares the generated production with the original brief, covering actor count, genre, theme alignment, content completeness and allowed lighting fixtures. If the first response fails either check, the implementation sends one correction request. This request contains the original requirements, machine-readable validation errors, the rejected output (treated as untrusted data) and the required schema. The corrected response is checked again and rejected if it still fails \cite{kumar2024llm}.

\subsection{Blocking and Cue Synchronization}
To keep movement references simple and machine-checkable, the stage is divided into nine zones: USL, USC, USR, CSL, CSC, CSR, DSL, DSC and DSR. A blocking cue records the actor, the starting zone, the destination zone, the action and the trigger.

\begin{figure}[H]
    \centering
    \includegraphics[width=0.84\columnwidth]{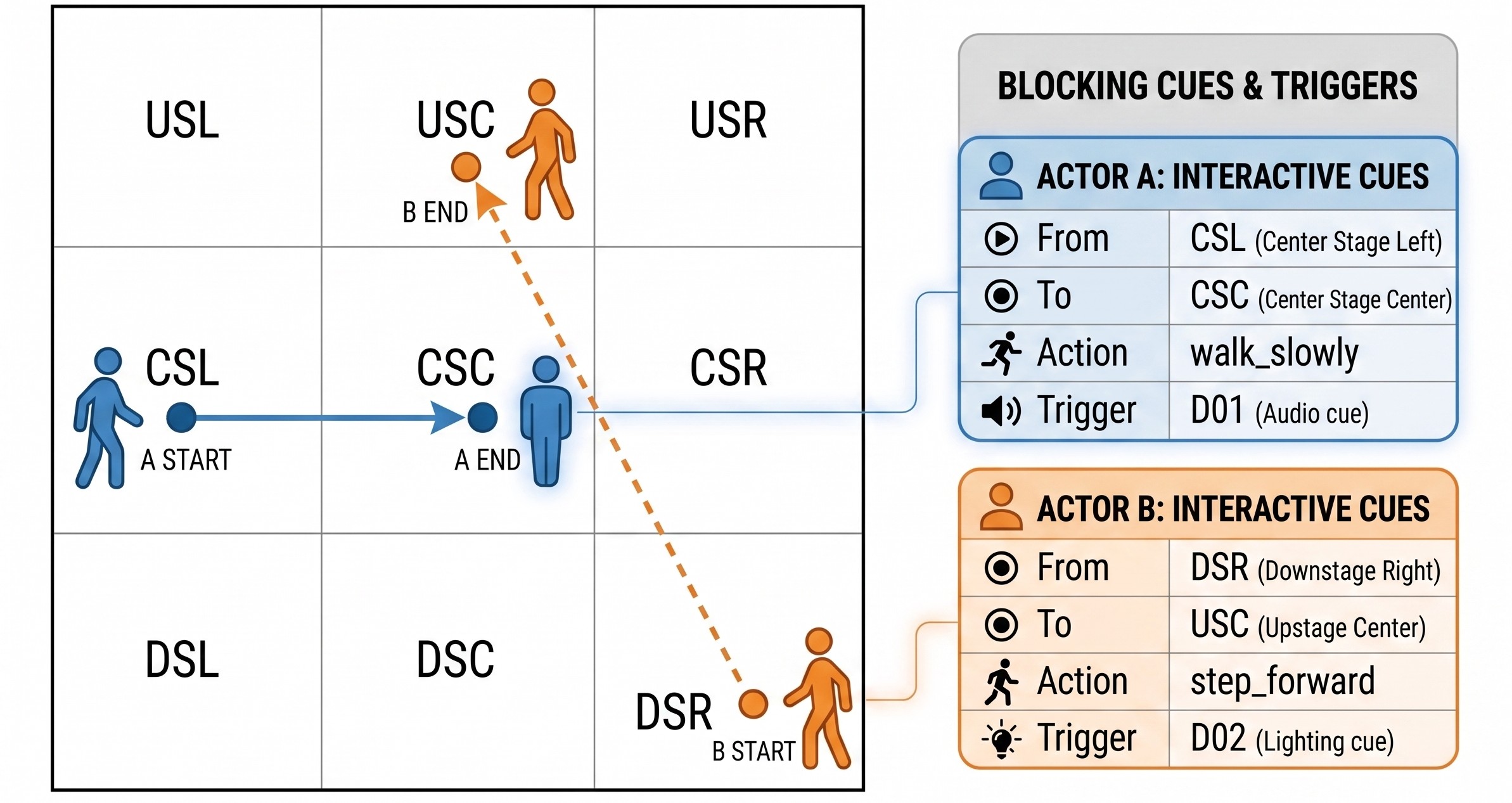}
    \caption{Nine-zone representation for actor blocking.}
    \label{fig:blocking}
\end{figure}

A lighting cue records its identifier, trigger, fixture, focus zone, RGB values, intensity, and fade duration. Dialogue, movement, lighting, and optional sound can therefore share the same event reference. Fig.~\ref{fig:blocking} shows movement as an explicit transition between known stage zones rather than ambiguous prose.

\begin{figure}[H]
    \centering
    \includegraphics[width=0.94\columnwidth]{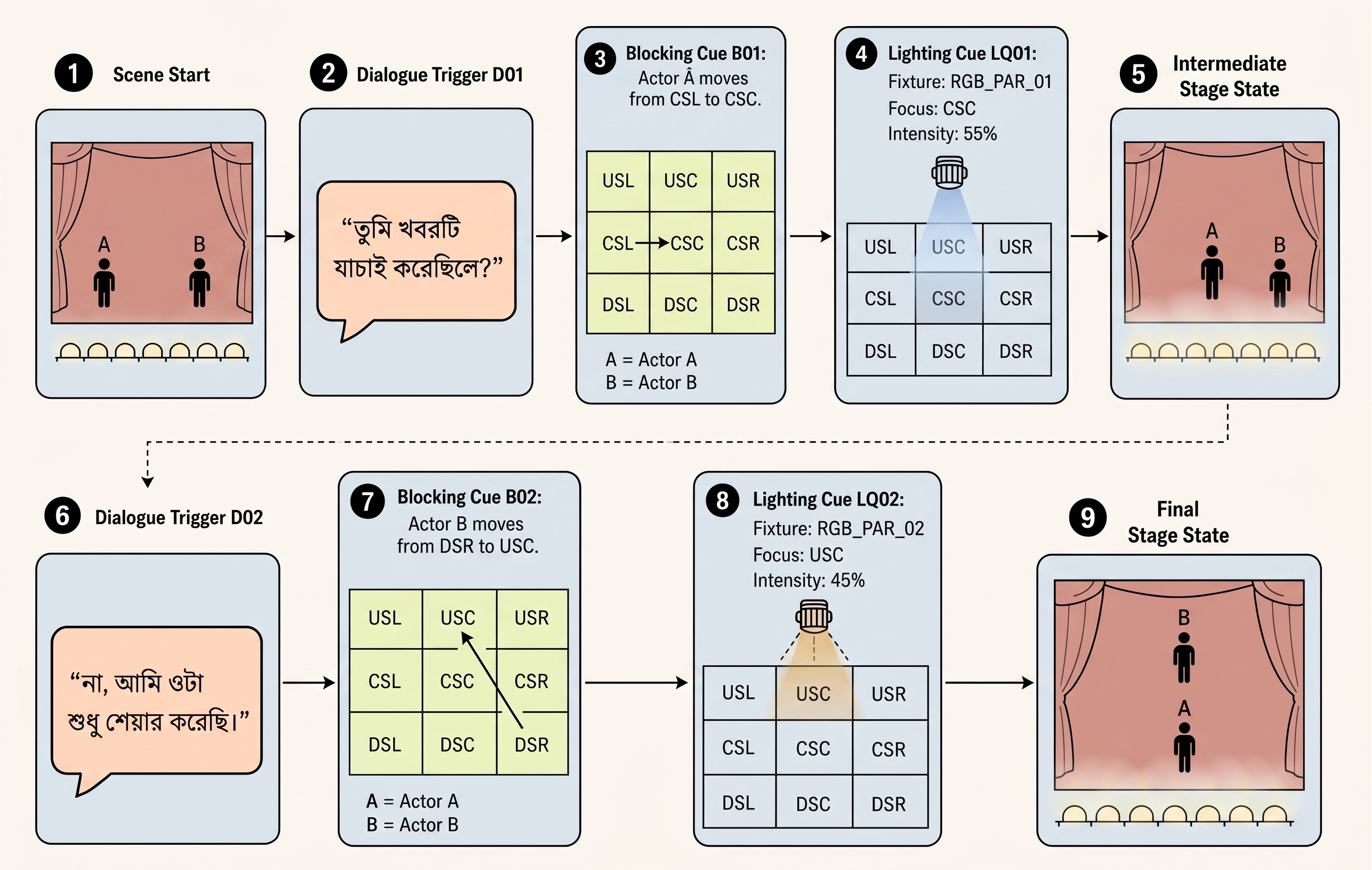}
    \caption{Event-aligned dialogue, blocking, and lighting cues.}
    \label{fig:synch}
\end{figure}

In Fig.~\ref{fig:synch}, a dialogue event is used as the common trigger for movement and lighting. The validator checks that the referenced event exists in the same scene. This event-driven organization is consistent with earlier work that links performance signals to stage lighting \cite{vryzas2017augmenting}.

\section{Bengali Theatre Dataset}
The current dataset was prepared as a compact synthetic test corpus for the retrieval and structured-generation workflow. It is not intended to represent the full range of Bengali theatre writing.

\subsection{Dataset Construction}
The 500 records were created through a human-guided, LLM-assisted process. LLM assistance was used to draft short records under the project schema, and the authors then reviewed and iteratively revised the structured fields, cross-references, dialogue, blocking, lighting, sound cues, and formatting used by the prototype. Each record contains an identifier, language, title, theme, genre, scene type, location, time, emotion, characters, dialogue, stage directions, actor blocking, lighting, sound cues, and a short summary. A common nested schema supports cross-reference validation, view-specific transformation, and provenance tracking. The corpus covers 15 themes, five genres, four scene types, eight locations, four times of day, five primary emotions, and 20 character names. The template-based synthetic construction also explains the high dialogue repetition reported in Section~VI.

\subsection{Multi-View Transformation}
\begin{figure}[H]
    \centering
    \includegraphics[width=0.7\columnwidth]{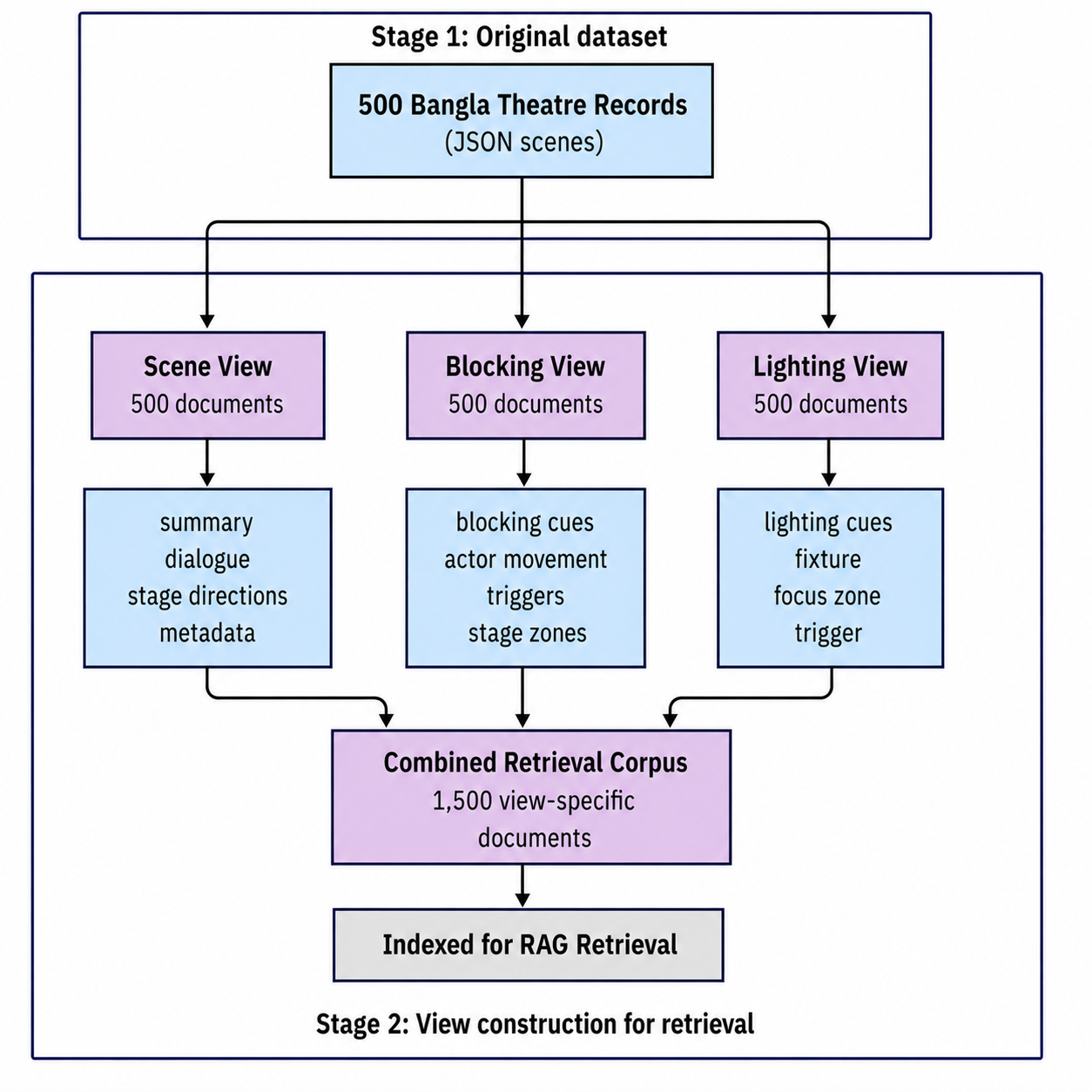}
    \caption{Transformation from source records to retrieval views.}
    \label{fig:datasetviews}
\end{figure}
From each source record, we create one scene document, one blocking document, and one lighting document. The scene payload keeps dialogue, stage directions, and summary information. The blocking payload keeps movement and stage-direction information. The lighting payload keeps lighting cues together with relevant blocking context. The source identifier is preserved in all three views.
As shown in Fig.~\ref{fig:datasetviews}, this one-to-three transformation converts 500 source records into 1,500 retrieval documents without losing provenance.

\subsection{Corpus Composition}
\begin{table}[H]
\caption{Dataset composition.}
\label{tab:dataset}
\centering
\footnotesize
\begin{tabular}{lr}
\toprule
\textbf{Component} & \textbf{Count} \\
\midrule
Original theatre records & 500 \\
Scene-view documents & 500 \\
Blocking-view documents & 500 \\
Lighting-view documents & 500 \\
\textbf{Total retrieval documents} & \textbf{1,500} \\
Dialogue lines & 3,000 \\
Blocking cues & 1,000 \\
Lighting cues & 1,000 \\
Sound cues & 1,000 \\
\bottomrule
\end{tabular}
\end{table}
The deterministic counts obtained from the released files are listed in Table~\ref{tab:dataset}. The three retrieval collections are equal in size, which makes later view-level comparisons easier to control. Equal size, however, should not be confused with equal retrieval difficulty.

\section{Implementation and Evaluation Setup}
We implemented the prototype as a set of services so that retrieval, generation, validation, and logging can be inspected separately rather than being buried inside the web interface.

\subsection{System Implementation}
The backend is built with Django. Separate service modules handle dataset loading, embedding, vector storage, retrieval, context construction, LLM access, validation, export, and experiment logging. This component-based structure follows common RAG engineering practice \cite{fan2024survey}. The main dependencies are Django, Pydantic, the official \texttt{google-genai} SDK, \texttt{sentence-transformers}, and \texttt{qdrant-client}. BGE-M3 produces multilingual embeddings, while local Qdrant storage maintains the scene, blocking, and lighting collections. Each generation run can store model settings, RAG mode, retrieved sources, retrieval trace, validation state, repair count, raw output, accepted structured output, and generation time. The present prototype stops at validated planning; it has not been deployed in a real theatre and does not directly actuate DMX lighting hardware.

\subsection{RAG Configurations}
Six retrieval modes are implemented: no RAG, scene-only RAG, scene plus blocking RAG, scene plus lighting RAG, single-combined retrieval, and full multi-view RAG. Full multi-view retrieval is the default, with top-$k$ set to 3 for scene, 2 for blocking, and 2 for lighting. The single-combined baseline retrieves at most seven documents so that its maximum reference count matches the multi-view setting. For the reviewer-requested comparison, 12 fixed Bengali production briefs are used in paired single-combined and full multi-view runs while keeping the Gemini model, generation settings, fixture constraints, retrieval budget, and validation policy fixed. This follows component-controlled RAG evaluation \cite{es2024ragas}.

\subsection{Evaluation Measures}
Generation is evaluated through validation pass rate, repair frequency, and end-to-end latency. The logs also retain scene, dialogue, blocking, and lighting counts. Retrieval measures such as Precision@$k$, Recall@$k$, MRR, and nDCG require relevance-labeled queries and are not inferred from the current unlabeled dataset \cite{saadfalcon2024ares}. Creative quality still requires human judgment, particularly for Bengali dialogue naturalness, stageability, blocking quality, lighting appropriateness, coherence, and cultural relevance \cite{chhun2024stories}.

\section{Results and Discussion}
The available evidence is reported as corpus diagnostics, retrieval-view characteristics, and generation results so that dataset properties are not mistaken for model performance.

\subsection{Corpus Integrity and Diversity}
The integrity checks found 500 unique record IDs and 500 unique titles. No duplicate flattened rows or missing flattened values were reported.

\begin{table}[H]
\caption{Selected corpus diagnostics.}
\label{tab:diagnostics}
\centering
\footnotesize
\setlength{\tabcolsep}{7pt}
\begin{tabular}{lr}
\toprule
\textbf{Diagnostic} & \textbf{Value} \\
\midrule
Unique record IDs & 500 / 500 \\
Unique titles & 500 / 500 \\
Rows with missing flattened values & 0 \\
Unique dialogue texts & 39 / 3,000 \\
Unique dialogue ratio & 1.3\% \\
Lighting cues with valid RGB values & 1,000 / 1,000 \\
Lighting cues with valid intensity & 1,000 / 1,000 \\
Lighting cues with non-negative fade & 1,000 / 1,000 \\
\bottomrule
\end{tabular}
\end{table}

All 1,000 lighting cues passed the recorded RGB-range, intensity-range, and non-negative fade-duration checks. The dialogue statistics show the main weakness of the current corpus: 3,000 lines contain only 39 distinct text strings, giving a 1.3\% unique-dialogue ratio. This repetition reflects the synthetic template-based construction and reduces the linguistic and dramatic variety available to retrieval and generation.

\subsection{Retrieval-View Characteristics}
Each retrieval collection contains 500 documents, and each view has 500 unique retrieval IDs and 500 unique source IDs. Across all three views, the diagnostics report 1,500 unique search texts.
\begin{figure}[H]
    \centering
    \includegraphics[width=0.8\columnwidth]{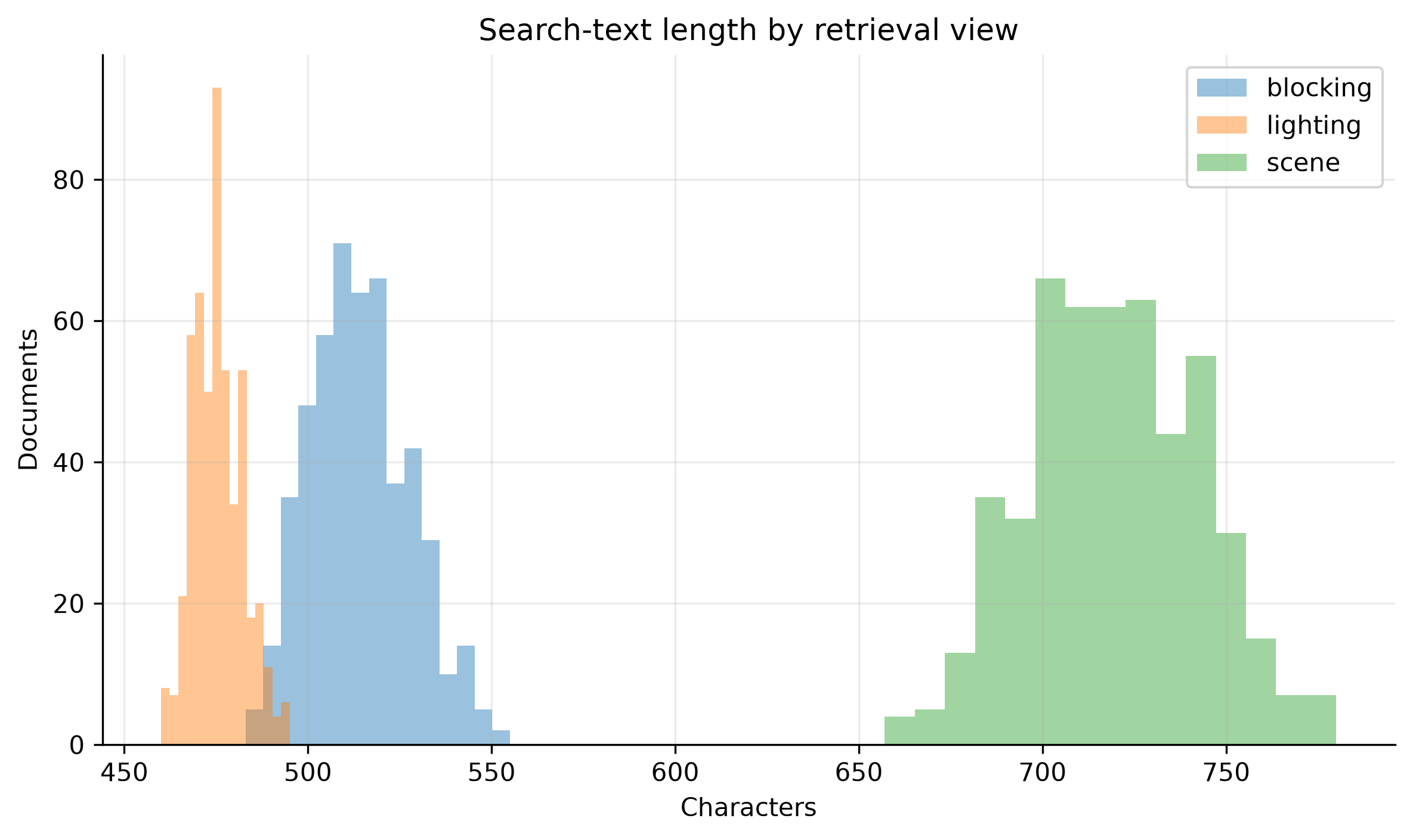}
    \caption{Search-text length distributions across retrieval views.}
    \label{fig:viewlengths}
\end{figure}
Mean search-text length is 719.3 characters for scene documents, 514.0 for blocking documents, and 475.4 for lighting documents. The scene view is longer because it retains more narrative material, whereas blocking and lighting views are more compact and technical. Fig.~\ref{fig:viewlengths} confirms that the three views expose different amounts of text to the retriever. It does not, by itself, establish that one view retrieves more relevant evidence than another. The blocking diagnostics also show that all nine stage zones occur as both source and destination zones.

\subsection{Prototype Execution Snapshot}
The original seven-run execution log is kept as a prototype diagnostic: five runs validated, two were invalid, the median generation time was 29.95 s and no repair attempt was recorded. Because this sample was small and uneven across modes, the camera-ready evaluation adds a paired comparison on 12 fixed briefs, run once with the single-combined baseline and once with the full multi-view configuration. This gives 24 runs under matched Gemini, validation, fixture and retrieval-budget settings.

\begin{table}[H]
\caption{Available prototype execution diagnostics.}
\label{tab:runs}
\centering
\footnotesize
\begin{tabular}{lccccc}
\toprule
\textbf{Setting} & \textbf{Runs} & \textbf{Valid} &
\textbf{Pass} & \textbf{Repair} & \textbf{Median s} \\
\midrule
All logged runs & 7 & 5 & 71.43\% & 0\% & 29.95 \\
Full multi-view & 4 & 4 & 100.0\% & 0\% & 30.55 \\
\bottomrule
\end{tabular}
\end{table}

The paired design keeps the reference budget and generation settings fixed, so it tells us more than the earlier seven-run snapshot. We release the fixed briefs and the evaluation script with the source code so the measurements can be reproduced. These automatic metrics assess constraint compliance and execution behavior, not artistic quality.

\subsection{Discussion and Limitations}
The results support the practical reason for keeping scene, blocking, and lighting retrieval separate. The views differ in textual size and remain independently traceable, so retrieval behavior can be inspected according to theatrical function. A scene reference that is useful for dialogue may not be the best source for actor movement or lighting.

The sustainability contribution is limited to resource-aware planning. Explicit fixture constraints and validation before physical execution can support more deliberate use of existing stage resources, but electricity consumption and environmental impact were not measured. The corpus contains only 500 short synthetic Bengali scripts and has high dialogue repetition; longer and more diverse scripts from theatre institutions, published plays, and other reliable sources are needed for stronger generalization. Lighting cues are generated and validated but have not been executed in a real theatre, which requires suitable instruments and DMX-compatible control hardware. Generation was tested only with Gemini using the available free-access service, so no cross-model conclusion is made. If a substantially larger and more diverse corpus becomes available, domain-specific fine-tuning can be explored alongside RAG. Structural validity also does not guarantee artistic quality, so human review remains necessary.

\section{Conclusion}
This work introduced ThetreStageAI, a prototype for retrieval-supported Bengali theatre production planning. Its central design choice is to retrieve scene, actor-blocking, and lighting material separately and bring those references together only before structured generation. Gemini produces a production object, and structural and semantic checks reject inconsistent actors, triggers, stage zones, fixtures, and other constraint violations. The human-guided, LLM-assisted synthetic test corpus contains 500 short records and 1,500 traceable retrieval documents. Diagnostics show strong structural consistency but limited dialogue diversity. The evaluation shows a paired comparison of the single-combined and full multi-view configurations under matched settings. The system is still a planning prototype. Broader conclusions would need a larger theatre corpus, human evaluation, comparison across models, and physical DMX lighting tests.

\section{Data and Code Availability}
The source code, the synthetic 500-record Bengali theatre dataset, the derived 1,500-document retrieval corpus and the camera-ready paired-evaluation materials are publicly available at \url{https://github.com/mehedinaeem/ThetreStageAI}.

\printbibliography

\end{document}